\documentclass[final,5p,times,twocolumn]{elsarticle}
\PassOptionsToPackage{hyphens}{url}
\usepackage{hyperref}
\usepackage{acronym}
\usepackage{xcolor}
\usepackage{soul}
\usepackage{amsmath}
\usepackage{graphicx}
\usepackage{placeins}
\usepackage{caption}
\usepackage{subcaption}
\usepackage{array}
\newcolumntype{C}{>{\centering\arraybackslash}p{1.2cm}}

\journal{Ecological Informatics}

\def\fref#1{Figure \ref{#1}}
\def\tref#1{Table \ref{#1}}
\def\secref#1{Section \ref{#1}}
\def\eqref#1{Eq. (\ref{#1})}\def\fref#1{Fig. \ref{#1}}
\def\eref#1{(\ref{#1})}

\acrodef{AR}{Augmented Reality}
\acrodef{AVAR}{Allan Variance}

\acrodef{COCO}{Common Objects in Context}
\acrodef{CRF}{Conditional Random Fields}

\acrodef{EU}{European Union}
\acrodef{Exif}{Exchangeable image file format}

\acrodef{GNSS}{Global Navigation Satellite System}
\acrodef{LUCAS}{Land Use/Cover Area frame Survey}

\acrodef{NSL}{Normalized Segment Length}

\acrodef{QGIS}{Quantum Geographic Information System}

\begin{document}
\begin{frontmatter}

\title{Skyline variations allow estimating distance to trees on landscape photos using semantic segmentation}
 


\author{Laura Martinez-Sanchez$^{1,*}$}
\author{Daniele Borio$^{1,*}$}
\author{Rapha\"el d'Andrimont$^{1}$}
\author{Marijn van der Velde$^{1}$}
\address{ $^{1}$\quad European Commission, Joint Research Centre (JRC), Ispra , Italy \\
	$^{*}$ These authors contributed equally to this work.}

\begin{abstract}
Approximate distance estimation can be used to determine fundamental landscape properties including complexity and openness. We show that variations in the skyline of landscape photos can be used to estimate distances to trees on the horizon. A methodology based on the variations of the skyline has been developed and used to investigate potential relationships with the distance to skyline objects.  
The skyline signal, defined by the skyline height expressed in pixels, was extracted for several \ac{LUCAS} landscape photos. Photos were semantically segmented with DeepLabV3+ trained with the \ac{COCO} dataset. This provided pixel-level classification of the objects forming the skyline. A \ac{CRF} algorithm was also applied to increase the details of the skyline signal. Three metrics, able to capture the skyline signal variations, were then considered for the analysis. These metrics shows a functional relationship with distance for the class of trees, whose contours have a fractal nature. In particular, regression analysis was performed against 475 ortho-photo based distance measurements, and, in the best case, a $R^2$ score equal to $0.47$ was achieved. This is an encouraging result which shows the potential of skyline variation metrics for inferring distance related information. 

\end{abstract}


\end{frontmatter}

\acresetall

\section{Introduction} 
\label{sec.intro}
\begin{figure*}[!ht]
	\centering
	\includegraphics[width=\linewidth]{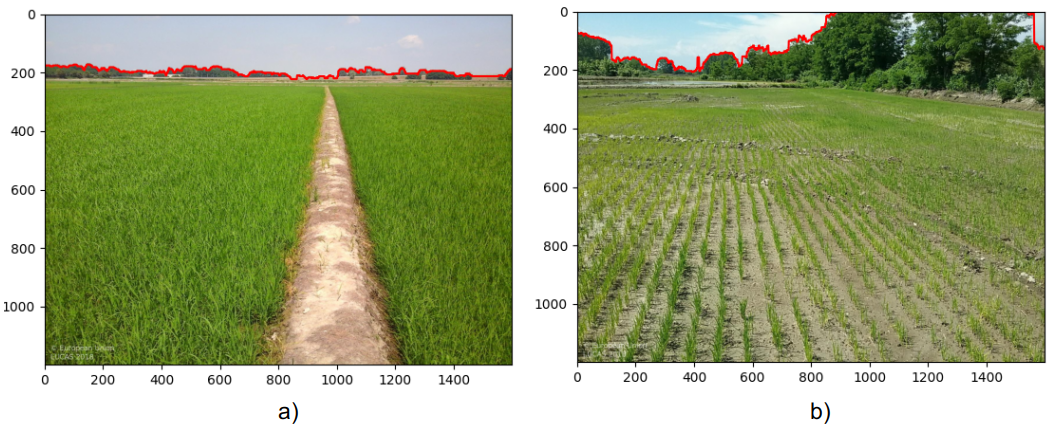}
	\caption{Comparison of skylines in red with background trees. a) Distant trees, b) Close trees.}
	\label{fig:principle}
\end{figure*}

The skyline, defined as the boundary between sky and non-sky (ground objects) of an image \cite{Johns2006,Ahmad2017}, provides significant information on the landscape and its complexity. Skyline profiles are used to evaluate the perception of urban landscapes by human subjects \cite{Ayadi2016} and can be adopted as indicators in \ac{AR} applications, for example to help visualize the impacts of new constructions \cite{Ayadi2018}. In this respect, several outdoor 
\ac{AR} applications requiring skyline extraction are emerging for low power mobile devices \cite{baboud2011automatic,frajberg2017convolutional,Touqeer2021skyline}. These applications are dedicated to fast identification of natural objects, such as plant species or mountain peaks.
Moreover, skyline identification is applied in mobile mapping systems where they are used as anchors to estimate camera orientations \cite{Hofmann2014}.
Skyline information can also assist the collection of in-situ land cover information, as done with e.g. the FotoQuest Go Europe campaign \cite{laso2020crowdsourcing}. Here, a horizontal line drawn across the screen assist the users to take photographs so that two-thirds of the photograph is land and one-third is sky.
\\ 
In addition to these applications, the shape of the skyline itself has been considered as a proxy to determine the (perceived) openness of a landscape \cite{Aifantopoulou2007, hedblom2020landscape}. The ratio between the length of the skyline and the width of the original image or the landform visual envelope contains such information \cite{Aifantopoulou2007}. 
In this paper, these ideas are developed further and several metrics computed from the skyline are proposed and related to the distances of the objects forming the skyline itself. The main goal is to establish empirical relationships between skyline metrics and object distances. These distances, in turn, can provide information on the landscape openness. A key point in the methodology developed is the detection and classification of the objects whose profiles from the skyline. This operation has been performed exploiting recent deep learning advances. In particular, semantic segmentation or pixel-wise classification was used to locate and delineate artificial and natural objects in photos \cite{taghanaki2021deep}. 
This operation was enabled by the availability of open image datasets which have pushed the evolution of semantic segmentation neural nets \cite{YU201882} which now achieve complex pixel-wise classification on both indoor and outdoor images.
\\ 
The idea developed in this study stems from the fact that skyline variations caused by similar objects strongly depend on the distance of such objects from the camera used to capture the image. For example, distant trees (\fref{fig:principle} a) cause only minimal variations in the skyline of an agricultural landscape while closer trees lead to more significant variations and discontinuities (\fref{fig:principle} b).
\\
This paper thus investigates the relationship between  object ground distance and skyline variations.The approach followed here is also inspired by the work of Mandelbrot \cite{Mandelbrot1983}. The classic example is that of a coastline that does not have a well-defined length. Since coastlines have fractal curve-like properties, the measured `length' depends on the scale and precision of the observation. Trees have similar properties. That is, the resolution of the roughness of a tree shape is impacted by its distance from the observer.   

The overall aim of this study is to identify metrics allowing to capture information on the distance to objects making up the skyline of a landscape photo. Specific objectives are: 
\begin{enumerate}
    \item To select a set of landscape photos from the \ac{LUCAS} survey in open agricultural landscapes. 
    \item To semantically segment the \ac{LUCAS} photos and retrieve their skyline. 
    \item To classify the objects making up the skyline and define metrics based on variations in the skyline. 
    \item To investigate relationships between skyline variation metrics and distances using independent distance measurements on aerial ortho-photos. 

\end{enumerate}

To achieve these goals, several variation metrics have been computed considering first order differences of the skyline signals. First order differences, which correspond to a numerical derivative, remove offsets in the skyline signal and operates as a high-pass filter which enhances high frequency signal variations. After computing first order differences, three metrics were evaluated: the normalized segment length, the sample variance, and the absolute deviation. Each metric has been computed considering skyline segments belonging to the same  object class. The impact of windowing, i.e. limiting the maximum length of the segment used for metric computation, has also been investigated.\\

The remainder of this paper is organized as follows: Materials and methods (\secref{sec.mame}) introduces the segmentation and classification methods adopted to extract the skyline signals and classify the different skyline objects, defines the evaluated distance metrics, and describes the gathering of reference distance data from aerial ortho-photos. The results of the distance-based analysis and the selection of the best metric are provided in \secref{sec.results}. Discussions and Conclusions are finally provided in \secref{sec.discussion} and \ref{sec.conclu}, respectively.

\section{Materials and methods} 
\label{sec.mame}
\subsection{Approach}
\begin{figure*}[!ht]
	\centering
	\includegraphics[width=0.8\linewidth]{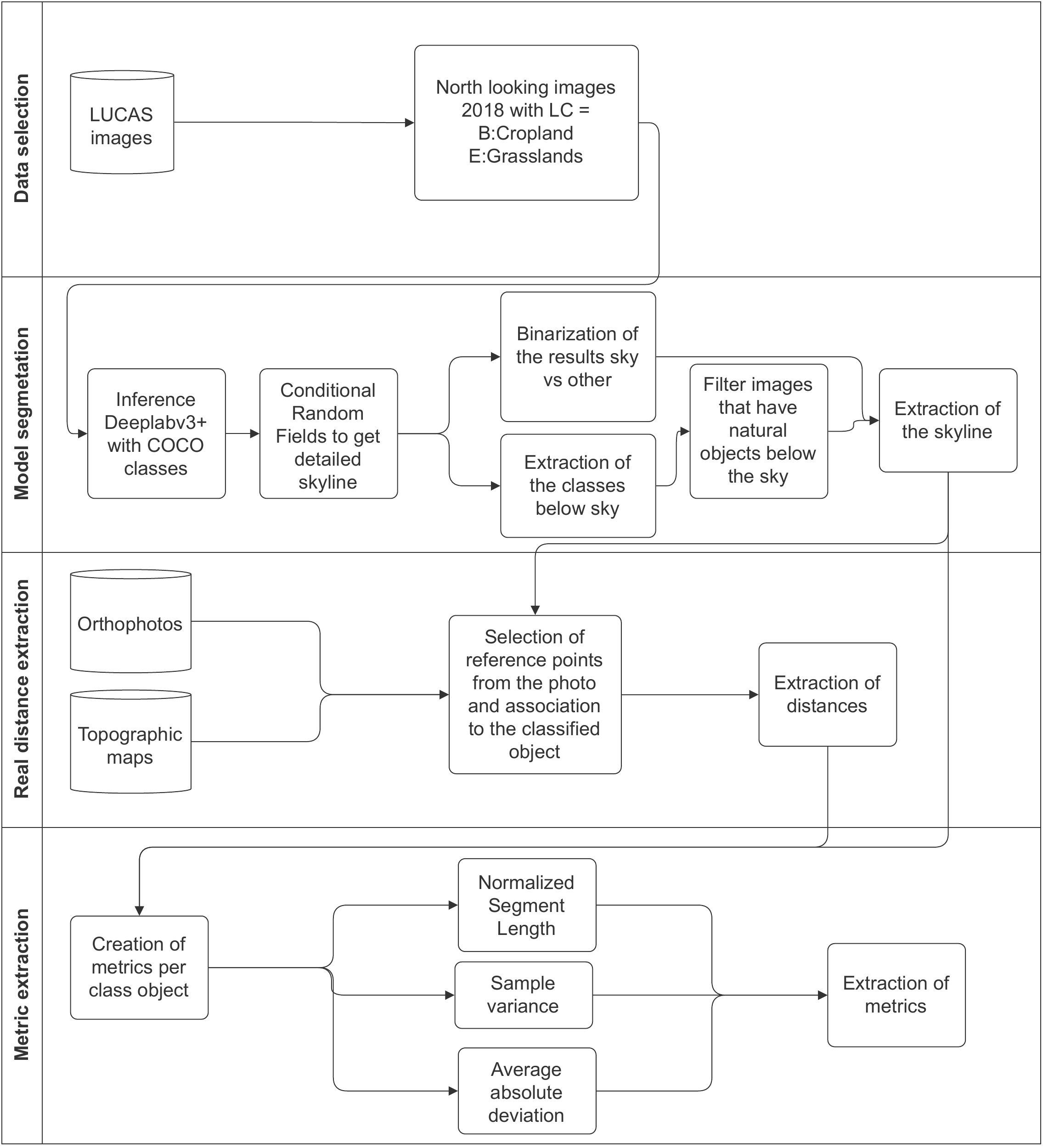}
	\caption{The workflow followed in this research.}
	\label{im:workflow}
\end{figure*}

Several methods have been proposed in the literature to retrieve the skyline of an image. These studies can be divided in three main groups, those based on traditional machine learning \cite{baatz2012, Yao2013, Saurer2015}, edge detection methods \cite{Lie2005,  Ayadi2016, jaciii2020}, and those based on deep learning methods \cite{ahmad2017comparison, badrinarayanan2017, Sassi2019}. In this work, skyline determination is performed using a semantic segmentation network trained with the \ac{COCO} dataset \cite{Lin2014},  discriminating sky versus other classes forming the skyline. 
After semantic segmentation, the objects that form the skyline are determined. Importantly, objects belonging to different classes will impact the shape of the skyline in different ways. For example, buildings with flat contours introduce different variations from those caused by natural objects like trees. These variations will scale differently as a function of distance. For this reason, the classifications of the pixels obtained by the semantic segmentation were used to identify the objects making up the skyline. Since we focus on open semi-natural landscapes, we considered trees, plants, houses and buildings as relevant classes from the \ac{COCO} dataset. Skyline variation metrics will be computed separately using pixels belonging to these classes. The hypothesis is that variations in the skyline profile of natural objects such as trees and plants will scale with the distance to the camera, while man-made objects such as buildings with flat profiles will not.  
\\
The first two steps of the analysis allow the identification of the skyline, which is treated as a discrete signal indexed with respect to the image horizontal pixel coordinate, $x$. Moreover, a second signal defining the object class is obtained for each value of the $x$ coordinate. These two signals are used for computing different candidate metrics that are finally analysed with respect to the object distance. Details are provided in the sections below and the workflow is shown in Figure \ref{im:workflow}.\\

\subsection{LUCAS landscape photos}
For the distance-based analysis, photos extracted from the \ac{LUCAS} 2018 \cite{d2020harmonised,rd2018} database were used. \ac{LUCAS} is a triennial in-situ land cover and land use data collection exercise that extends over the whole of the \ac{EU}’s territory (\url{https://ec.europa.eu/eurostat/web/lucas}). 
\ac{LUCAS} collects information on land cover and land use, agro-environmental variables, soil, and grassland.  At each \ac{LUCAS} point, standard variables are collected including land cover, land use, environmental parameters (the so called \textit{micro data}), as well as one downward facing photo of the point (P) and four landscape photos in the cardinal compass directions (N, E, S, W). Additionally, each photo has attached \ac{Exif} attributes such as coordinates, orientation, camera model, exact time and date, Eurostat metadata, etc. 
\subsubsection{LUCAS landscape photo selection}
A set of $148$ North looking \ac{LUCAS} photos with different and easily identifiable skyline objects were selected in this study. These photos were obtained by randomly selecting from the \ac{LUCAS} land cover classes B 'croplands' and  E 'grasslands'. After running the inference with Deeplabv3+, we filter out photos where the sky was not present on at least the 10 upper rows, thus eliminating photos where the view was obstructed. A second filter ensured that photos were selected where the tree class was present just below the skyline. 

\subsection{Semantic segmentation}
\label{ssec.semseg}
\begin{figure*}[!ht]
    \centering
	\includegraphics[width=0.7\linewidth]{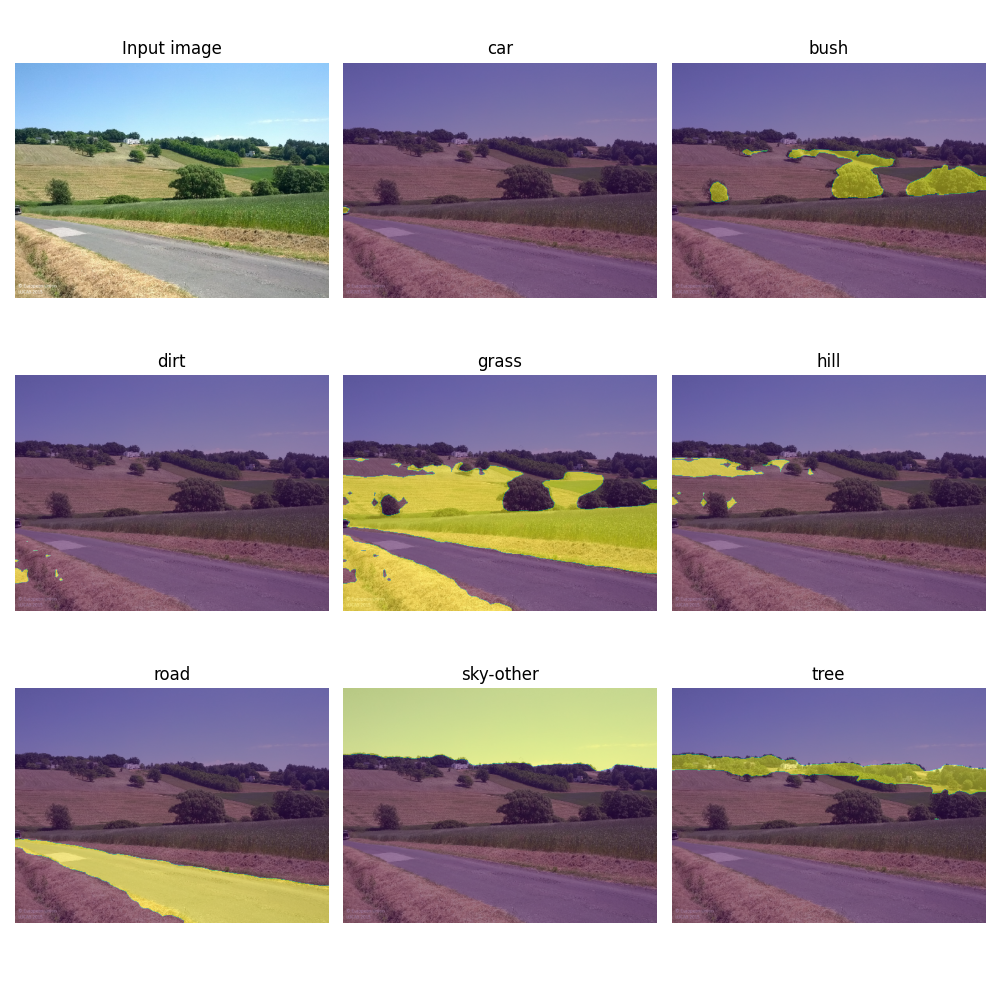}
	\caption{Semantic segmentation of a LUCAS landscape photo with DeeplabV3+. This example shows the input image (upper left) and output masks corresponding to COCO classes: car, bush, dirt, grass, hill, road, sky-other and tree.}
	\label{im:deeplabout}
\end{figure*}

The skyline was extracted using a semantic segmentation net trained with the \ac{COCO} dataset. The net used is DeeplabV3+, an encoder/decoder neural net that improves upon DeepLabv2 by using atrous convolution to handle the problem of segmenting objects at multiple scales \cite{Chen2018}. In order to take advantage of transfer learning, the weights were obtained from an already trained implementation of DeeplabV3+ (\url{https://github.com/zllrunning/deeplab-pytorch-crf}) with the \ac{COCO} dataset. \ac{COCO} is a dataset that has $121,408$ images and a total of $883,331$ object annotations divided in $184$ classes. \ac{COCO} is a widely used dataset to train computer vision models and sets a baseline to benchmark their performance. Since we want to derive common objects below the skyline, we inference without training the model directly on the landscape photos. A typical example of the masks obtained with such a semantic segmentation is provided in Figure \ref{im:deeplabout}. 

To extract meaningful information for the purpose of our study, the delineation of the skyline has to be highly detailed. The detection of edges or object boundaries in semantic segmentation tends to be blurry as the loss of effective spatial resolution associated with the learning of contextual information by the net diminishes high-frequency details \cite{Shelhamer2017, hariharan2015hypercolumns}. We address this problem by applying a \ac{CRF}\cite{krahenbuhl2011efficient}. The ability to capture fine details with a \ac{CRF} compensates the limited performance of these kind of nets to generate precisely delineated objects.

\begin{figure*}[!ht]	
\centering
	\includegraphics[width=0.8\linewidth]{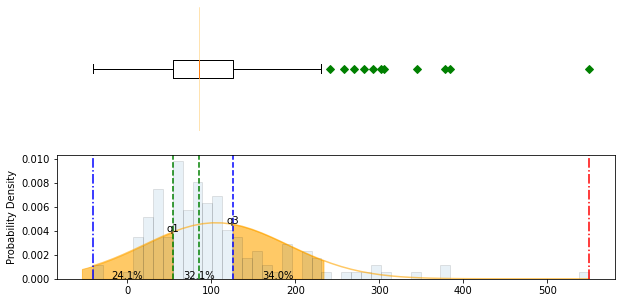}
	\caption{Distribution of the normalized length differences of the skyline signals obtained with and without \ac{CRF}. In average, the \ac{CRF} increases the signal length by more than 100\%. The bar plot and the histogram show the actual distribution of the gain with \ac{CRF}, showing also the quantiles (dotted lines).}
	\label{fig:DensityCRF}
\end{figure*}

To measure the increase in detail of the skyline using \ac{CRF}, we calculated the length of the signals extracted with and without \ac{CRF} and evaluated the gain length as:
\begin{equation}
    g_{CRF} = \frac{\mbox{len}(y_{sky}[x])}{\mbox{len}(y_{sky}^{noCRF}[x])} - 1
\end{equation}
where $y_{sky}[x]$ is the skyline signal obtained with \ac{CRF} and $y_{sky}^{noCRF}[x]$ is determined without. $\mbox{len}(\cdot)$ is the total length of the signal computed using \eqref{eq:norm_length} and extending the summation to the whole image width. $g_{CRF}$ can also be interpreted as the normalized length difference between the skyline signals obtained with and without \ac{CRF}. A negative gain implies a reduction of the signal length. \fref{fig:DensityCRF} provides the histogram and the statistics of length gain provided by the \ac{CRF}.

\subsection{Reference distance measurements on ortho-photos}
\label{ssec.refdata}
\begin{figure*}[!ht]
    \centering
    \includegraphics[width=0.7\linewidth]{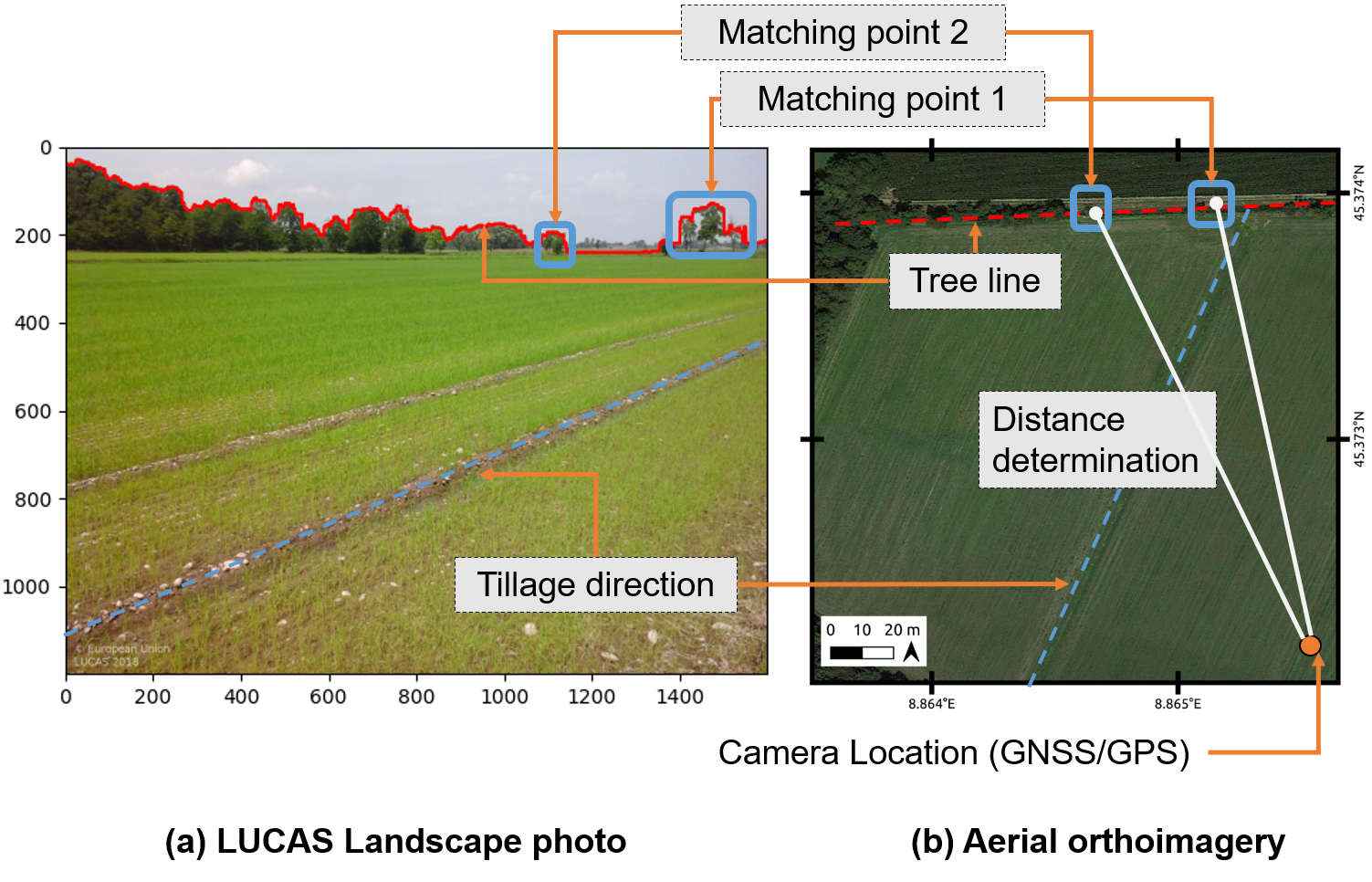}
    \caption{Illustration of the process adopted to measure distances from the camera to skyline objects using aerial photos. The background RGB ortho-imagery is obtained from “Map data \textcopyright 2021 Google”.}
    \label{fig:ref_distance}
\end{figure*}

As described above, a reference dataset was built considering photos from the \ac{LUCAS} database. Each \ac{LUCAS} photo contains the camera location as part of its metadata. This information was extracted and plotted in a \ac{QGIS} project along with aerial photos extracted from open aerial or satellite photo databases such as Google Satellite and Bing Aerial. For each \ac{LUCAS} photo, reference points from the skyline were identified in the corresponding aerial photo.
This process is illustrated in \fref{fig:ref_distance}: at first, common elements such as tillage direction and tree lines were identified in both \ac{LUCAS} photo (a) and aerial images (b). These elements allow to properly frame the \ac{LUCAS} photo with respect to the aerial image and simplify the process of identifying reference points present in the skyline of the \ac{LUCAS} photo.
In \fref{fig:ref_distance}, an isolated tree is at first considered on the right of the \ac{LUCAS} photo. This point is easily found in the aerial image following the tree line. In this case, the tillage direction also simplifies the identification of this point.\\
Depending on the image considered, different elements were determined and used for identifying reference skyline points.
For each \ac{LUCAS} photo, several skyline points were considered and identified in the corresponding aerial image. Finally, for each point, the distance from the camera location was determined using the measuring tools available in \ac{QGIS}. Only \ac{LUCAS} photos for which a reliable identification of skyline reference points was possible were considered. A set of $150$ photos was selected including different classes of skyline objects. On average, three reference points were mapped for each image with a total set of $485$ elements.\\
For the $i$th location point, the following information was recorded:
\begin{itemize}
    \item \textbf{POINT ID}: identifier of the \ac{LUCAS} point and thus photo.
    \item \textbf{Point coordinates}: $(x_i, y_i)$ coordinates in pixels in the original \ac{LUCAS} photo.
    \item \textbf{Distance}, $d_i$, from the reference point to the camera.
\end{itemize}
Note that the measured distances are affected by uncertainties, which arise from difficulties in precisely identifying the location of reference points and by errors in the measuring process. A significant effort was made to reduce these residual uncertainties by excluding photos where no clear point was identifiable in the skyline and by using additional data such as topographic maps (OpenTopoMap\url{http://www.opentopomap.org}). The residual errors present in the measured distances are considered sufficiently small not to compromise the analysis detailed in the following.    

\subsection{Skyline signal and distance metrics}
Our hypothesis is based on two premises. The first is that objects that are closer to the position of the camera will have more pixels representing them compared to objects that are far away. The second is that natural objects such as trees have fractal like shapes that lead to skyline signals that scale with distance. This implies a higher frequency in variations of the skyline profile of trees that are close, compared to those that are farther away.\\

\begin{figure*}[!ht]
	\centering
	\includegraphics[width=0.6\linewidth]{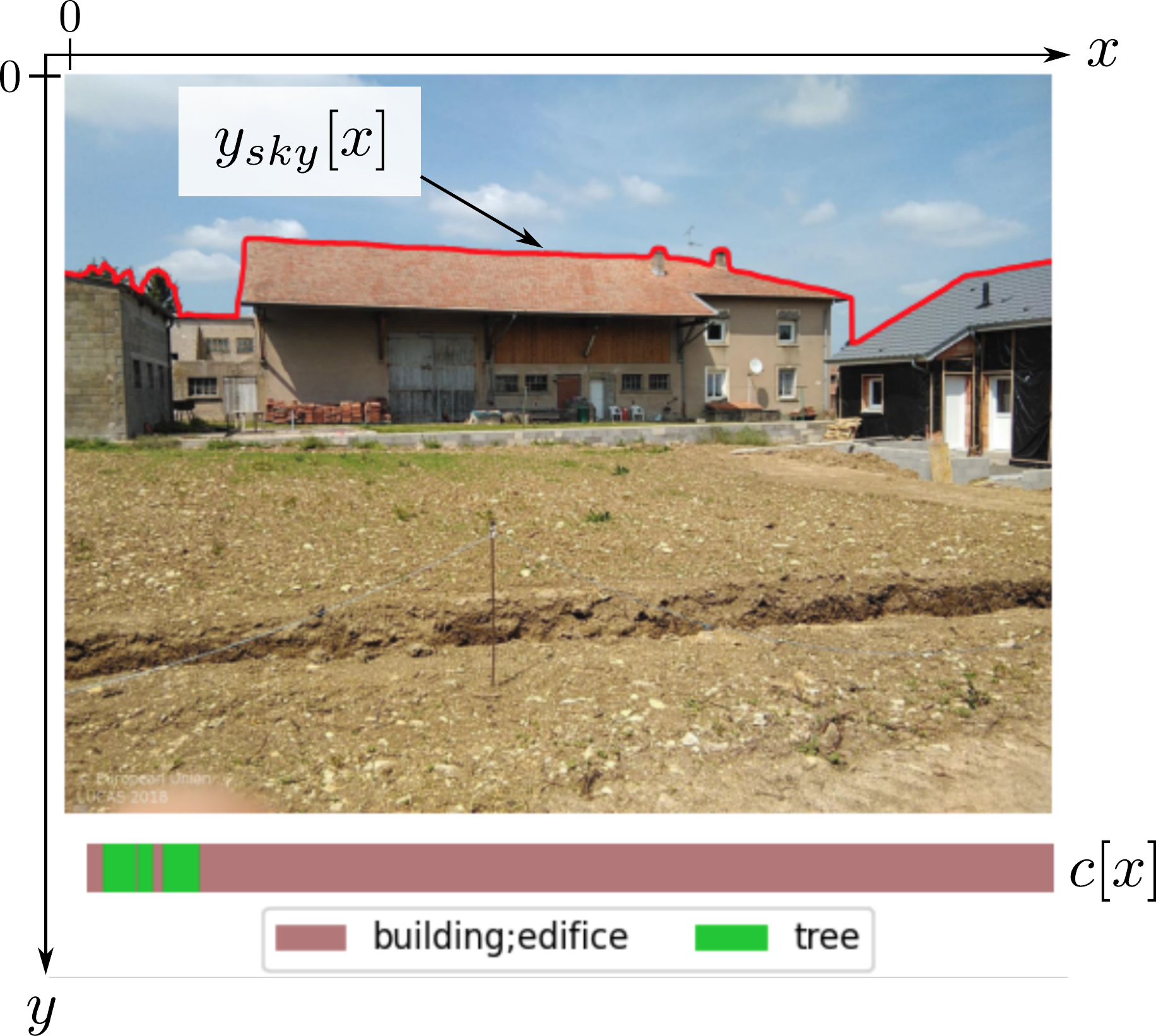}
	\caption{The skyline and classification signals extracted from a \acs{LUCAS} photo.}
	\label{fig:signals}
\end{figure*}

The output of the semantic segmentation process is the discrete signal:
\begin{equation}
	y_{sky}[x] \qquad \mbox{for } 0 \leq x < W,
\label{eq:sky_sig}
\end{equation}
which takes values in the set, $\{0, H\}$, where $W$ and $H$ are the width and height of the original image. For each $x$ value, $y_{sky}[x]$ is the height of the skyline in the image. 
The second signal indicates to which class pixel $(x, y_{sky}[x])$ belongs. This is expressed as:
\begin{equation}
	c[x] = \sum_{k = 0}^{Nx-1}c_k\Pi_{D_k}[x - x_{c,k}]
	\label{eq:class_sig}
\end{equation}
where $c_k$ is the $k$th class value and $\Pi_{D_k}[x - x_{c,k}]$ is the casual unit door defined as
\begin{equation}
	\Pi_{D_k}[x - x_{c,k}] = \left\{
		\begin{array}{lr}
		1 & \mbox{for } x_{c,k} \leq x < x_{c,k} + D_k\\
		0 & \mbox{otherwise}.
		\end{array}
	\right.
	\label{eq:unit_door}
\end{equation}
$N_x$ is the number of objects defining the skyline. In this way, $c[x]$ is piece-wise constant on intervals of type, $[x_{c,k}, x_{c,k} + D_k]$. Each interval defines a skyline segment for which an object belonging to class $c_k$ has been identified.\\
These signals are shown in \fref{fig:signals}: $c[x]$ is depicted at the bottom of the figure as a bar where each color represents a different class. 
From this representation the piece-wise nature of the signal clearly appears.\\
In several images, it has been observed that the skyline is affected by variations not due to the shape of the skyline objects but due to geometric effects. For example, a line of trees can be slanted because it is placed along a sloped hill that introduces horizontal variations in addition to changes due to the shapes of the trees. These geometric variations can be significant and hide the actual effects of the objects. For this reason, first-order differences have been computed:
\begin{equation}
\dot{y}_{sky}[x] = y_{sky}[x] - y_{sky}[x-1] \quad \mbox{for } 1 \leq x < W.
\label{eq:first_order}
\end{equation}
By performing first-order differences, which is approximately equivalent to compute a derivative, all linear trends present in $y_{sky}[x]$ are removed.
$\dot{y}_{sky}[x]$ and $c[x]$ are the basic signals used in the following for the computation of metrics quantifying the variability of the skyline.
\subsection{Variability metrics}
\label{ssec.metrics}
As discussed in \secref{ssec.refdata}, a data set of $485$ points with associated distances to objects as measured on ortho-photos has been created. For each point, identified by the the coordinates $(x_i, y_i)$ with $y_i = y_{sky}[x_i]$, we compute a metric, $m_i$, to quantify the variability of the corresponding skyline profile. In order to do so, the class of the point is at first identified using \eqref{eq:class_sig}. This class is denoted as $c_i = c[x_i]$. The boundaries of the corresponding skyline object are also determined as 
\begin{equation}
\begin{split}
    x_{c,i} &= \max\left\{x_{c,k} \leq x_i\right\}\\
    D_{i} &= x_{c,{i+1}} - x_{c,i}\\
    \label{eq:point_boundaries}
\end{split}
\end{equation}
where $x_{c,k}$ and $D_k$ have been defined above. $x_{c,i}$ and $D_i$ defines the range of points belonging to the same class of $x_i$: only these points are used for the metric computation. Several metrics are considered in the following. These metrics have been selected since they are typical indicators used in the literature to quantify the variability of a signal.
\subsubsection{\ac{NSL}}
This is the original metric proposed by \cite{Aifantopoulou2007} and adapted here to individual objects forming the skyline. This metric is computed for the $i$th object as:
\begin{equation}
\begin{split}
m_{nsl,i} &= \frac{1}{D_i - 1}\sum_{x_{c,i} + 1}^{x_{c,i} + D_i - 1}\sqrt{(y_{sky}[x] - y_{sky}[x-1])^2 + 1^2} \\
		  &= \frac{1}{D_i - 1}\sum_{x_{c,i} + 1}^{x_{c,i} + D_i - 1}\sqrt{\dot{y}_{sky}[x]^2 + 1}.
\end{split}
\label{eq:norm_length}
\end{equation}
The total length is computed as the sum of the lengths of the individual segments obtained by considering subsequent pixels. The two terms under square root are the 
horizontal and vertical variations between consecutive pixels. For this reason, the horizontal variation is always equal to $1$. $m_{nsl,i}$ depends only on $\dot{y}_{sky}[x]$ and is normalized by $D_i - 1$, the length that would be obtained if $\dot{y}_{sky}[x] = 0$ for all $x$.
\subsubsection{Sample variance}
The second metric considered for the analysis is the sample variance of $\dot{y}_{sky}[x]$ defined as \cite{Casella2001}:
\begin{equation}
    m_{sv,i} = \frac{1}{D_i - 1}\sum_{x = x_{c,i} + 1}^{x_{c,i} + D_i - 1}\left(\dot{y}_{sky}[x] - \overline{\dot{y}_{sky}}\right)^2
    \label{eq:variance}
\end{equation}
where $\overline{\dot{y}_{sky}}$ is the sample mean:
\begin{equation}
    \overline{\dot{y}_{sky}} = \frac{1}{D_i-1}\sum_{x = x_{c,i} + 1}^{x_{c,i} + D_i - 1}\dot{y}_{sky}[x].
    \label{eq:sample_mean}
\end{equation}
Note that both sample variance and mean are computed considering samples in the interval $[x_{c,i}, x_{c,i} + D_{i} - 1]$, i.e. using samples belonging to the same object class. Moreover,
\begin{equation}
	\begin{split}
\overline{\dot{y}_{sky}} &= \frac{1}{D_i-1}\sum_{x = x_{c,i} + 1}^{x_{c,i} + D_i - 1}\left(y_{sky}[x] - y_{sky}[x-1]\right)\\
&= \frac{1}{D_i-1}\left[\sum_{x = x_{c,i} + 1}^{x_{c,i} + D_i - 1}y_{sky}[x] - \sum_{x = x_{c,i}}^{x_{c,i} + D_i - 2}y_{sky}[x]\right]\\
&= \frac{1}{D_i-1}\left(y_{sky}[x_{c,i}+D_i-1] - y_{sky}[x_{c,i}]\right) 
	\end{split}
\label{eq:sample_mean_2}
\end{equation}
For large values of $D_i$ and small signal variations, $\overline{\dot{y}_{sky}} \approx 0$ and
\begin{equation}
	m_{sv,i} \approx \frac{1}{D_i - 1}\sum_{x = x_{c,i} + 1}^{x_{c,i} + D_i - 1}\left(y_{sky}[x] - y_{sky}[x-1]\right)^2
	\label{eq:variance_2}
\end{equation}
which is a form of \ac{AVAR} \cite{Allan1966}.
\subsection{Average absolute deviation}
\eqref{eq:variance_2} suggests others variability metrics where different powers of signal differences are considered. In particular, \eqref{eq:variance_2} can be generalized as
\begin{equation}
m_{p,i} = \frac{1}{D_i - 1}\sum_{x = x_{c,i} + 1}^{x_{c,i} + D_i - 1}\left|y_{sky}[x] - y_{sky}[x-1]\right|^p = \frac{1}{D_i - 1}\sum_{x = x_{c,i} + 1}^{x_{c,i} + D_i - 1}\left|\dot{y}_{sky}[x]\right|^p
\label{eq:power_p}
\end{equation}
While we have analysed different values of $p$, in the following only the case $p = 1$ is analysed. This value led to the most promising results along with the sample variance. For $p =1$, $m_{1,i}$ represents the average absolute deviation of $\dot{y}_{sky}[x]$. 
\subsubsection{Windowed metrics}
\label{sssec.wm}
\begin{figure*}[!ht]
	\centering
	\includegraphics[width=0.8\columnwidth]{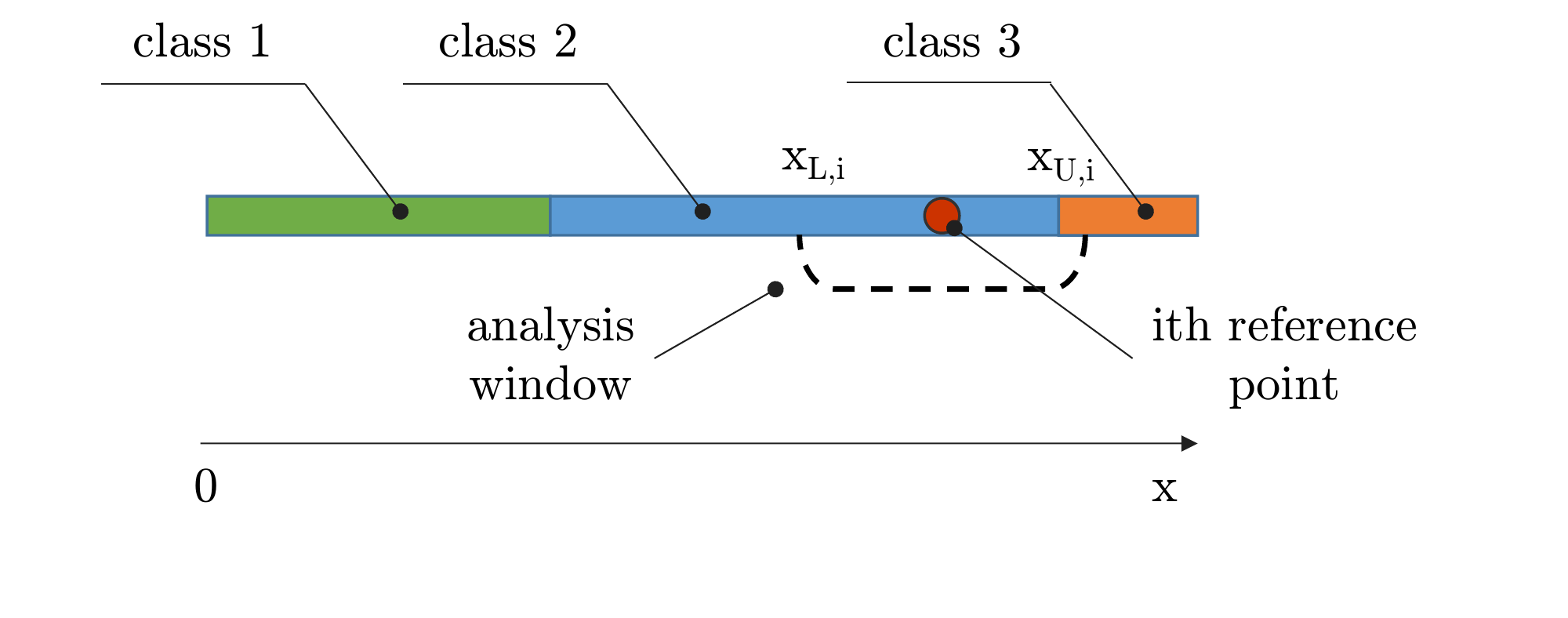}
	\caption{Schematic representation of the windowing process adopted to generalize dispersion skyline metrics.}
	\label{fig:windowing}
\end{figure*}

The metrics described above can be interpreted as averages computed considering a whole segment over which a single object class was detected. Windowed version of such metrics can be computed by limiting the duration of the summations in the metric computation. The windowing process adopted to generalize the skyline metrics is illustrated in \fref{fig:windowing}.
In this case, the different metrics are computed as:
\begin{equation}
m_{i}(L) = \frac{1}{x_{U,i} - x_{L,i} - 1}\sum_{x = x_{L,i} + 1}^{x_{U,i} - 1}f\left(\dot{y}_{sky}[x]\right)
\label{eq:windowed_metrics}
\end{equation}
where
\begin{itemize}
	\item $L$ is the length of the analysis window
	\item 
	\[
		x_{L,i} = \max\left(x_{c,i}, x_i - \lfloor L/2\rfloor \right)
	\]
	\item
	\[
		x_{U,i} = \min\left(x_{c,i} + D_i, x_i + \lfloor L / 2\rfloor \right)
	\]
	\item $f(\cdot)$ is used to denote the generic function used for the metric computation. For example, for the metric in \eqref{eq:power_p}, $f(\cdot) = |\cdot|^p$.
\end{itemize}
As previously defined, $x_i$ is the horizontal coordinate of the reference point for which the metric is computed.\\
Windowing should improve the ability of the metrics in mapping the distance to skyline objects. The rationale of applying windowing is to limit the effects of objects at different distances in the metrics computation. Consider for example, a skyline made of a line of trees. These trees could be aligned in a slanted way and have different distances from the camera. Without windowing, all the skyline points of this line of trees will be used for the metric computation since they all belong to the same class (trees). However, a hypothetical reference point $x_i$ refers to a single tree with a specific distance from the camera. Windowing reduces the impact of contamination of the nearby trees in the metric computation selecting a maximum of $L$ points around $x_i$.\\
Windowing also reduces the occurrence of duplicated metric values. Consider the case where two reference points in the same image belong to the same object. This could happen for example when both a chimney and the top of a roof are selected when considering the same house profile. Since the two points belong to the same object, they will lead to the very same metric value if windowing is not applied. In such a case, two different distance values will be associated to the same metric. Windowing is thus introduced also to reduce this effect. 
~\\
In the following, the three metrics described above and their windowed versions will be described. While other metrics were considered, they are not discussed since they did not show a dependency with respect to the camera distance.     
\subsection{Metric-to-distance models}
\label{ssec.models}
The metrics discussed above have been computed using the reference data generated according to the procedure discussed in \secref{ssec.refdata} and plotted as a function of the measured distances, $d_i$. In this way, scatter plots have been obtained. When both $m_i$ and $d_i$ are plotted in logarithmic scale, a linear trend was observed. In the following, symbol $m_i$ is used to denote a generic metric: the appropriate subscript is used to indicate specific quantities.\\
While the scatter plots obtained for the different metrics will be discussed in the ``Results'' section, the linear trends observed when logarithmic scales are adopted suggested the use of linear models for regression analysis \cite{Casella2001,Draper1998}:
\begin{equation}
\log(m_i) = \alpha + \beta\log(d_i).
\label{eq:log_linmodel}
\end{equation}
This model implies a power law between $m_i$ and $d_i$,
\begin{equation}
m_i = \mbox{e}^\alpha d_i^\beta = k\cdot d_i^\beta.
\label{eq:powerlaw_model}
\end{equation}
In the following, regression analysis using model \eref{eq:log_linmodel} is performed. The goodness of the regression is assessed using the coefficient of determination, $R^2$ \cite{Casella2001,Draper1998}.\\
Where appropriate, the following simplified model will also be considered in addition to \eqref{eq:powerlaw_model}:
\begin{equation}
m_i = \frac{k}{d_i}.
\label{eq:inverse_power}
\end{equation}
This model implies that the level of the variations (length) or fractal curve increases with the inverse of distance.

\FloatBarrier
\section{Results}
\label{sec.results}
\subsection{Semantic segmentation}

\begin{figure*}[!ht]
     \centering
        \begin{subfigure}[b]{0.5\linewidth}
            \centering
            \includegraphics[width=\linewidth]{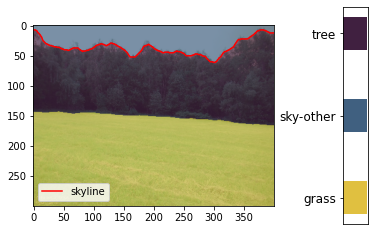} 
            \caption{Without CRF} 
            \vspace{4ex}
        \end{subfigure}
        \begin{subfigure}[b]{0.5\linewidth}
            \centering
            \includegraphics[width=\linewidth]{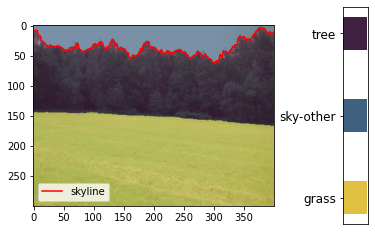} 
            \caption{With CRF} 
            \vspace{4ex}
        \end{subfigure} 
    \caption{Difference in the detail of the skyline signal obtained using  DeepabV3+ without (a) and with \acs{CRF} (b).}
    \label{fig:CRF-default}
\end{figure*}

The application of the \ac{CRF} improves the accuracy and detail of the skyline determination significantly, as illustrated in \fref{fig:CRF-default}, and explained in \secref{ssec.semseg}.The \ac{CRF} was run for a total of $15$ iterations on each photo, and the skyline was generated as a signal by simple post-processing that separated the sky from the other objects
From the figure, it emerges that on average the \ac{CRF} doubles skyline signal length. More specifically, the skyline length increased in average by 105.47\%, with a median gain of 85.53\%.
This confirms that the \ac{CRF} effectively increases the level of detail in the skyline signal. 

\subsection{Evaluating distance metrics}
In this section, results obtained by considering the different metrics are provided. Given the reasoning laid out before, we consider individual trees to be specifically well suited objects to derive distance and to evaluate the computed metrics. For this reason,  the large majority of objects selected during the generation of the reference data belongs to the class of ``Trees''. Trees are also more easily identifiable in the aerial photos with respect to other objects. Nonetheless, we also obtained significant sets for the other object types (``Houses'', ``Other Plants'' and ``Other Buildings''). All four classes were used to test the hypothesis and considered for the regression analysis. The distribution of reference points with respect to the different object classes is provided in \tref{tab:object_distr}.
\begin{table*}
	\centering
	\caption{Distribution of the reference points as a function of the different object classes.}
	\label{tab:object_distr}
\begin{tabular}{|C|C|C|C|C|C|C|}
	\hline
	\textbf{Object class}& Trees  & Houses  & Other Plants & Other Buildings & Other Classes &  Total \\
	\hline
	\textbf{No. of Points}&  300  & 44 & 33 & 46 & 62 & 485\\
	\hline
\end{tabular}
\end{table*}
 \\

\begin{figure*}[!ht]	
	\centering
	\includegraphics[width=0.7\linewidth]{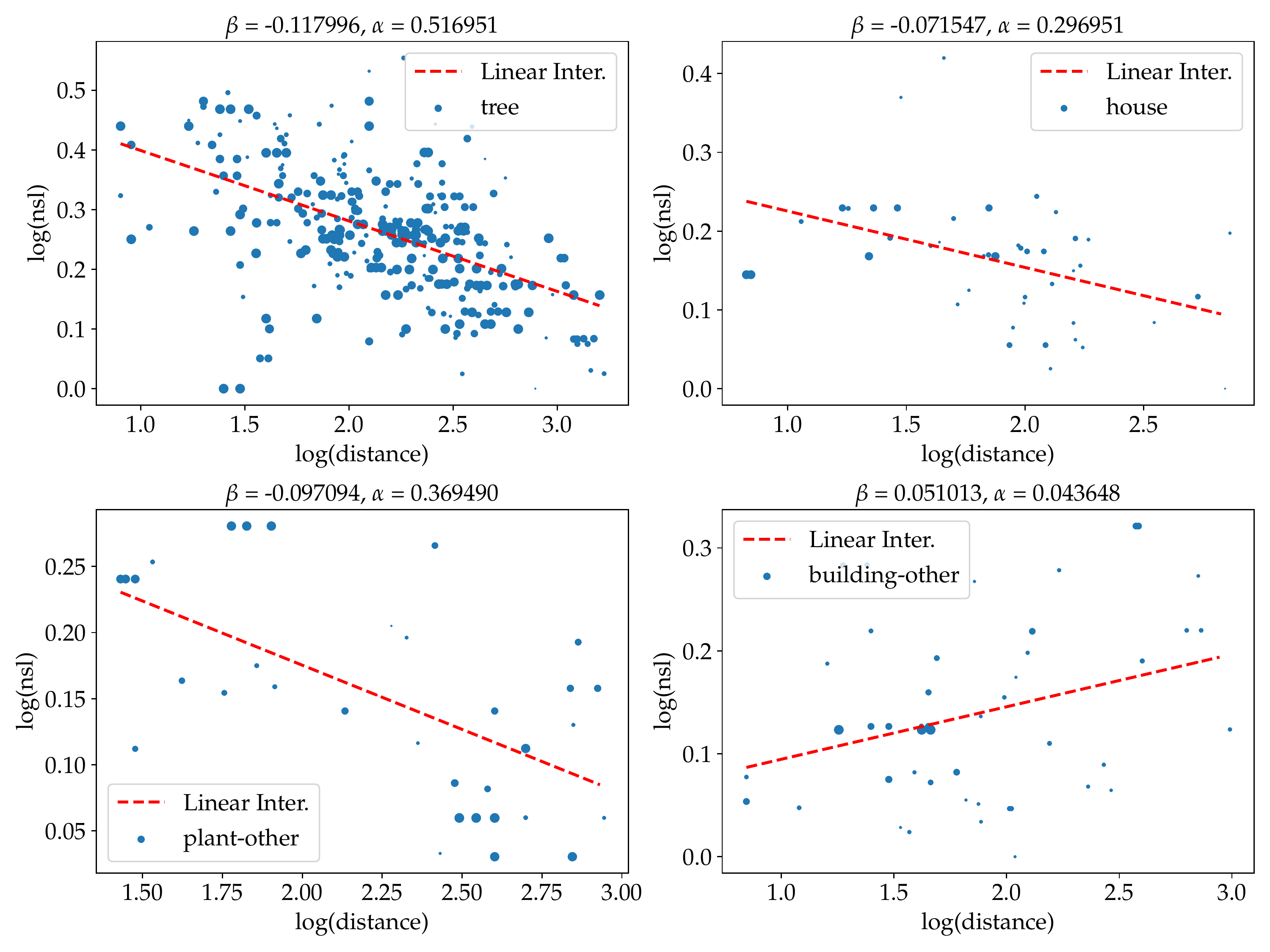}
	\caption{Scatter plots of the \ac{NSL} as a function of the estimated distance for different object classes. Regression lines are also provided.}
	\label{fig:normlengthreg}
\end{figure*}

\begin{figure*}[!ht]	
	\centering
	\includegraphics[width=0.7\linewidth]{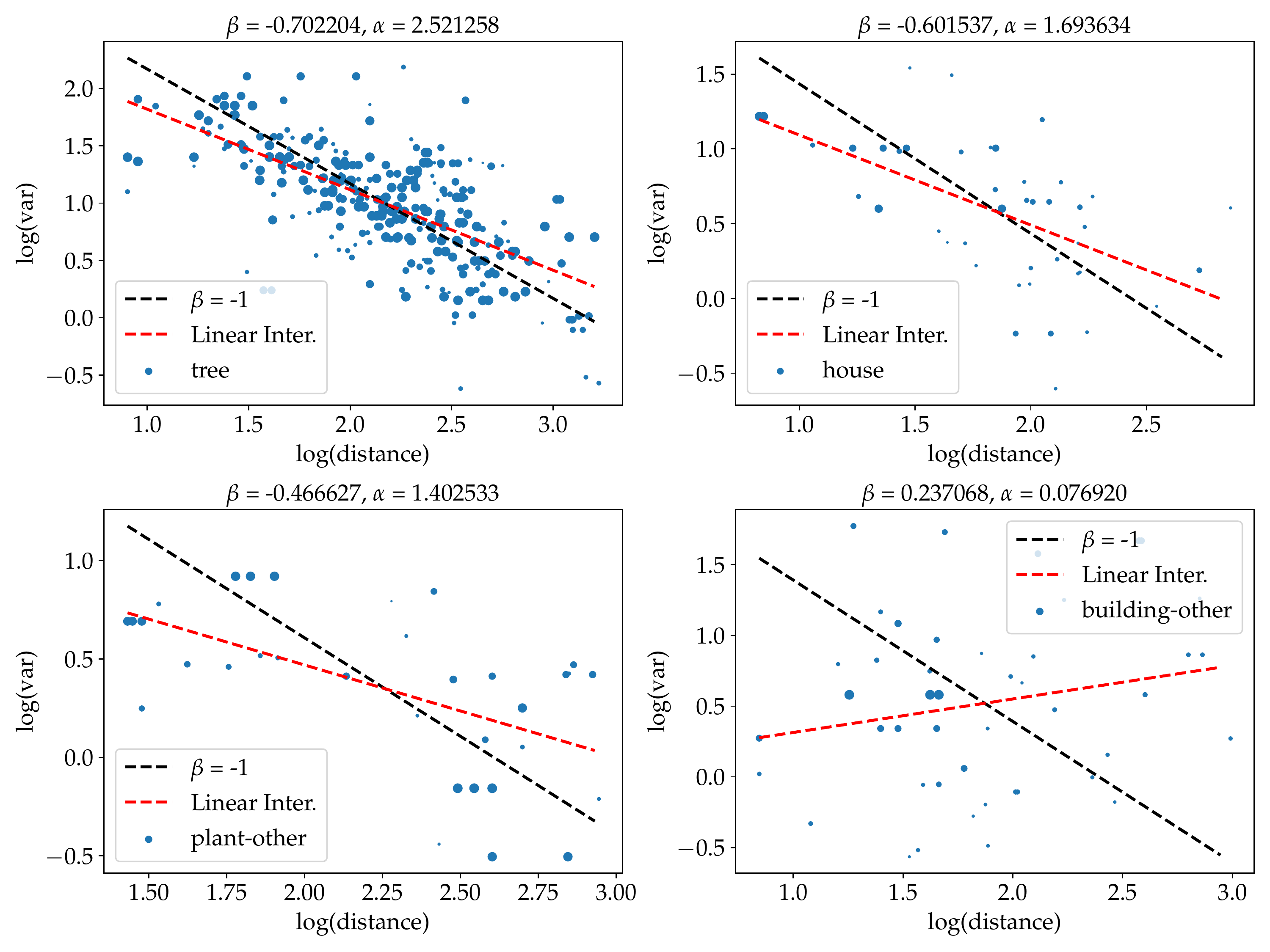}
	\caption{Scatter plots of the sample variance as a function of the estimated distance for different object classes. Regression lines are also provided along with the lines obtained for the constrained model with $\beta = -1$.}
	\label{fig:sample_var_reg}
\end{figure*}

\begin{figure*}[!ht]	
	\centering
	\includegraphics[width=0.7\linewidth]{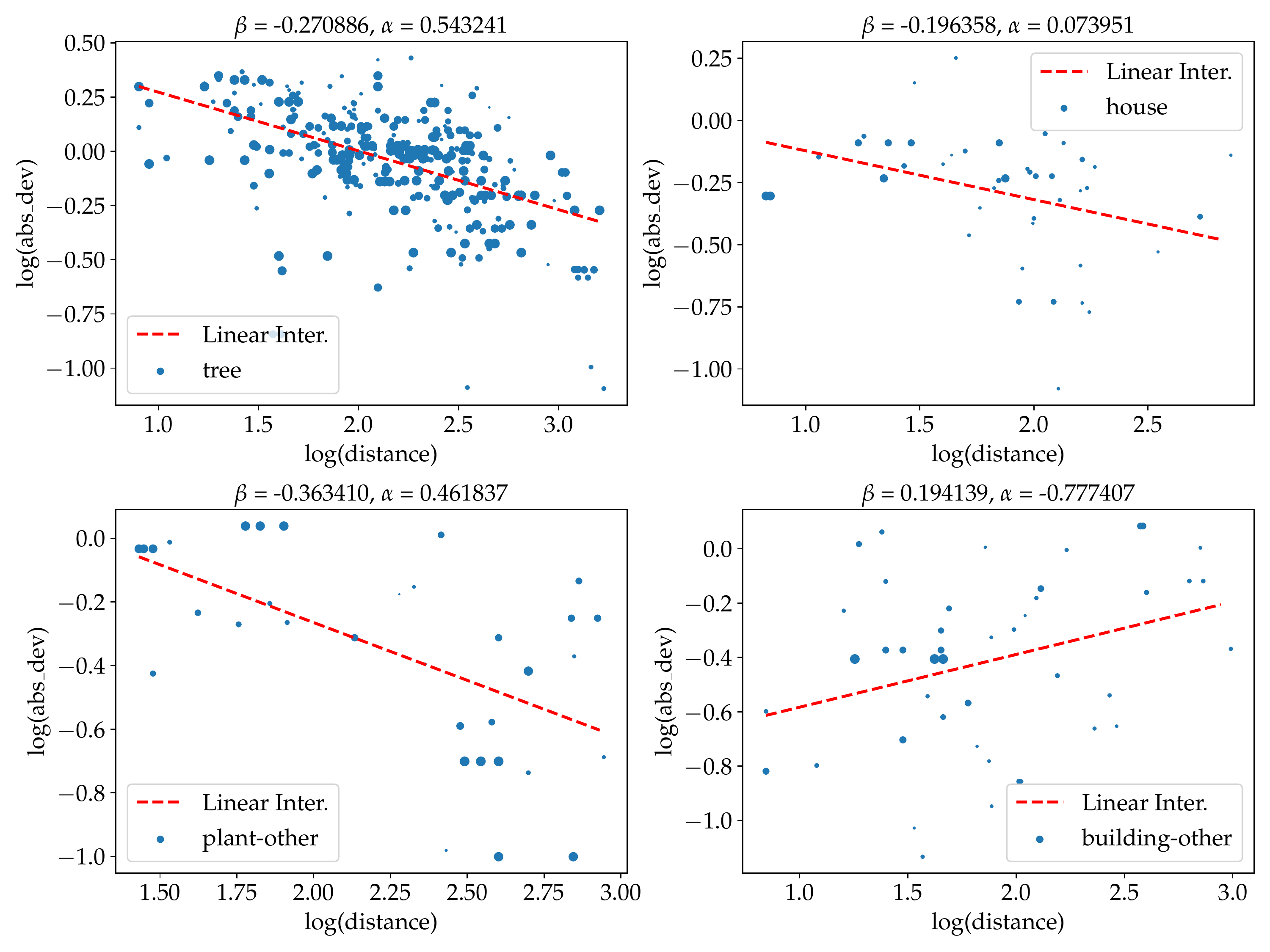}
	\caption{Scatter plots of the absolute deviation as a function of the estimated distance for different object classes. Regression lines are also provided.}
	\label{fig:abs_deviation_reg}
\end{figure*}

\begin{figure*}[!ht]	
	\centering
	\includegraphics[width=0.7\linewidth]{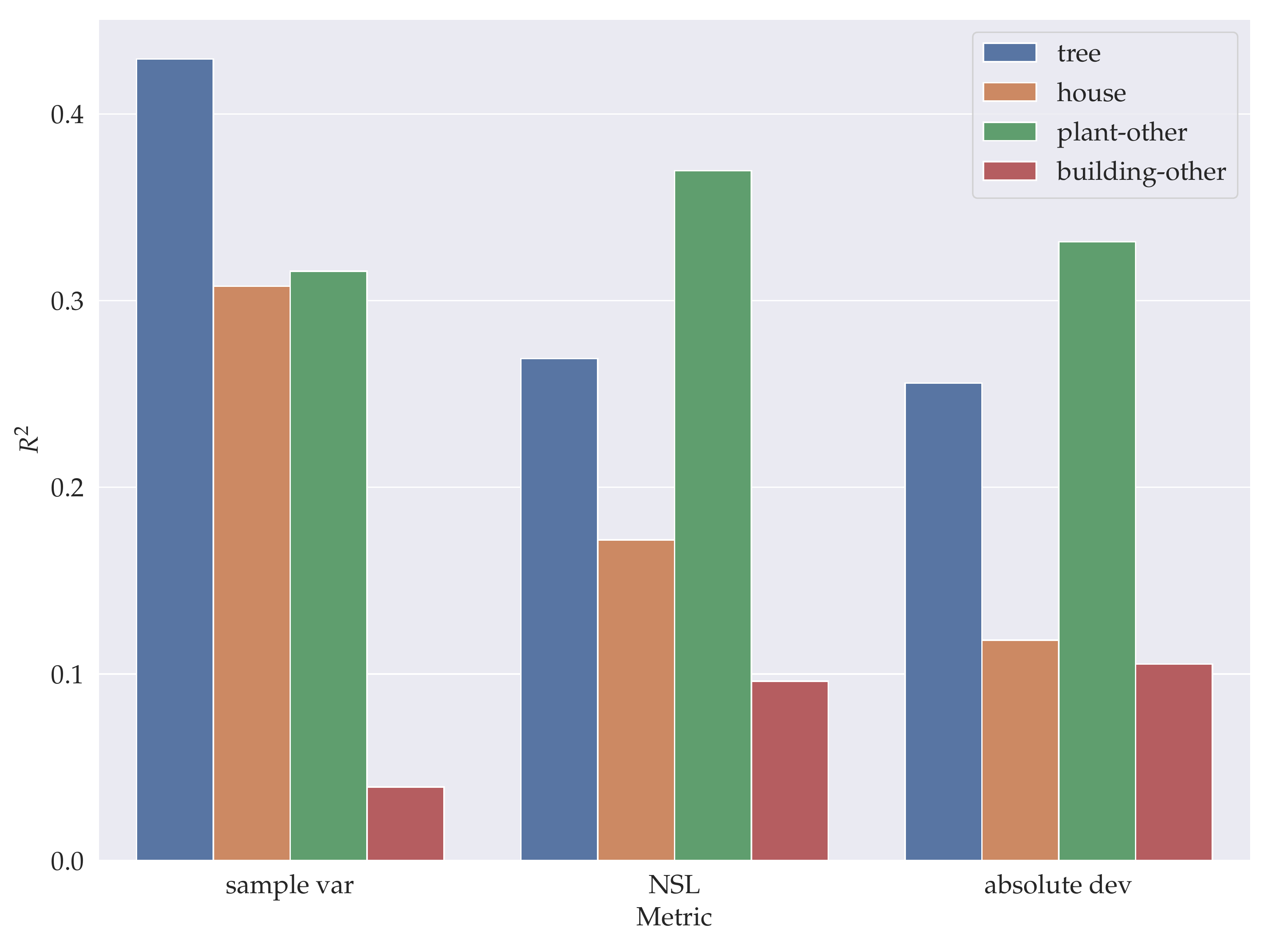}
	\caption{Comparison of the $R^2$ values obtained for the different metrics and object classes.}
	\label{fig:r2valuescomp}
\end{figure*}

In  \tref{tab:object_distr}, only classes with more than $30$ points are reported. Among the $485$ points analysed, $62$ belongs to other $8$ classes which are not analysed in the following given their reduced size in terms of reference points." 
Metrics without windowing are analysed at first. In particular, the scatter plots of the \ac{NSL} are provided in \fref{fig:normlengthreg} as a function of distance and for the four classes listed above. As already discussed both metrics and distances are expressed in logarithmic scale.   
For the ``Trees'' class, the scatter plot of the \ac{NSL} metric is clearly elongated along the regression line indicated in red. In this case, a coefficient of determination, $R^2 = 0.27$ was found. For three classes (``Trees'', ``Houses'' and ``Other Plants''), the slope coefficient, $\beta$, is negative indicating a decrease of \ac{NSL} with distance. In all four cases, however, $\beta$ assumes values lower than $0.12$ in magnitude. A larger value for $\beta$ is obtained for the class of ``Trees''. These results suggest that the \ac{NSL} metric is only weakly dependent on the distance, $d_i$.\\
A stronger dependence is observed for the sample variance metric whose scatter plots are shown in \fref{fig:sample_var_reg}. In this case, larger values, in magnitude, of $\beta$ are found and, in particular, $\beta = -0.7$ is obtained for ``Trees''. As for the previous case, $\beta$ is negative for all classes but ``Other Buildings''.\\
\fref{fig:sample_var_reg} also shows the regression lines obtained by fixing $\beta = -1$: in this case only $\alpha$ is estimated as
\begin{equation}
\alpha = \frac{1}{N}\sum_{i=0}^{N-1}(\log(m_i) + \log(d_i)).
\label{eq:alpha} 
\end{equation}
The lines obtained for $\beta = -1$ are depicted in black: for the case of ``Trees'' this line is close to the regression curve and it is able to explain a significant portion of the variability observable in the scatter plot. This fact is better analysed in \tref{tab.R2varComp} that provides the $R^2$ values for the different cases.  
When linear regression is adopted, almost $43\%$ of the variability is explained by the model for the ``Trees'' class. This value is slightly reduced to about $35\%$ when model \eref{eq:inverse_power} is considered: these results indicate the inverse power dependency between the sample variance and object distance for the class of trees. Lower values of $R^2$ are found for the other classes. Note that for ``Houses'' and ``Other Buildings'' no dependency was expected. This expectation was induced by the fact that most of the skyline profiles found for these two object categories were made of straight segments which look similar at different distances. This expectation has been confirmed by the class of ``Other Buildings'' which is characterized by low or negative $R^2$ values. The positive $R^2$ values found in \tref{tab.R2varComp} for the class of ``Houses'' are better analysed in the following when the effect of windowing is discussed. In particular, the profiles belonging to this class are analysed and used to justify the results obtained in \tref{tab.R2varComp}.   
\begin{table*}
	\centering
	\caption{Comparison of the $R^2$ values obtained for the sample variance considering linear regression and the constrained model with $\beta = -1$.}
	\label{tab.R2varComp}
	\begin{tabular}{|c|c|c|c|c|}
		\hline
		& Trees & Houses & Other Plants  & Other Buildings \\
		\hline
		Linear Model & 0.4294 & 0.3077 & 0.3158 & 0.0394 \\
		\hline
		Fixed Slope ($\beta = -1$ ) & 0.3522 & 0.1727 & -0.0968 & -1.034 \\
		\hline
	\end{tabular}
\end{table*}
It is also important to keep in mind that a lower number of reference points was used for the classes of ``Houses'', ``Other Plants'' and ``Other Buildings'': for such classes the results obtained should be considered less reliable than in the case of ``Trees''.\\
Finally, the scatter plots obtained for the absolute deviation metric are provided in \fref{fig:abs_deviation_reg}.
Also in  this case, an elongated scatter plot is found for the class of ``Trees'' where the regression model achieves an $R^2 = 0.255$. The absolute deviation shows a behaviour similar to that observed for the \ac{NSL} where reduced values of $\beta$ are found. As for the previous cases, the regression curve has a positive slope only for the ``Other Buildings'' class, for which however a low $R^2$ value is found.\\
The $R^2$ values obtained for the different metrics and object classes are compared in \fref{fig:r2valuescomp}. As already discussed, the sample variance is the metric leading to the largest $R^2$ for all classes but ``Other Buildings'' which is the class leading to the lowest $R^2$ values. ``Trees'' and ``Other Plants'' always lead to $R^2$ larger than $0.25$.  
The results in \fref{fig:r2valuescomp} confirm that model \eref{eq:powerlaw_model} explains a significant portion of the variability observed in the skyline metrics. The sample variance shows the most significant dependency on distance for natural objects such as trees and other plants. As expected, man-made objects such as ``Houses'' and ``Other Buildings'' show only a marginal dependency on the metric.\\

%

\subsection{Windowing}
\begin{figure*}[!ht]	
	\centering
	\includegraphics[width=0.7\linewidth]{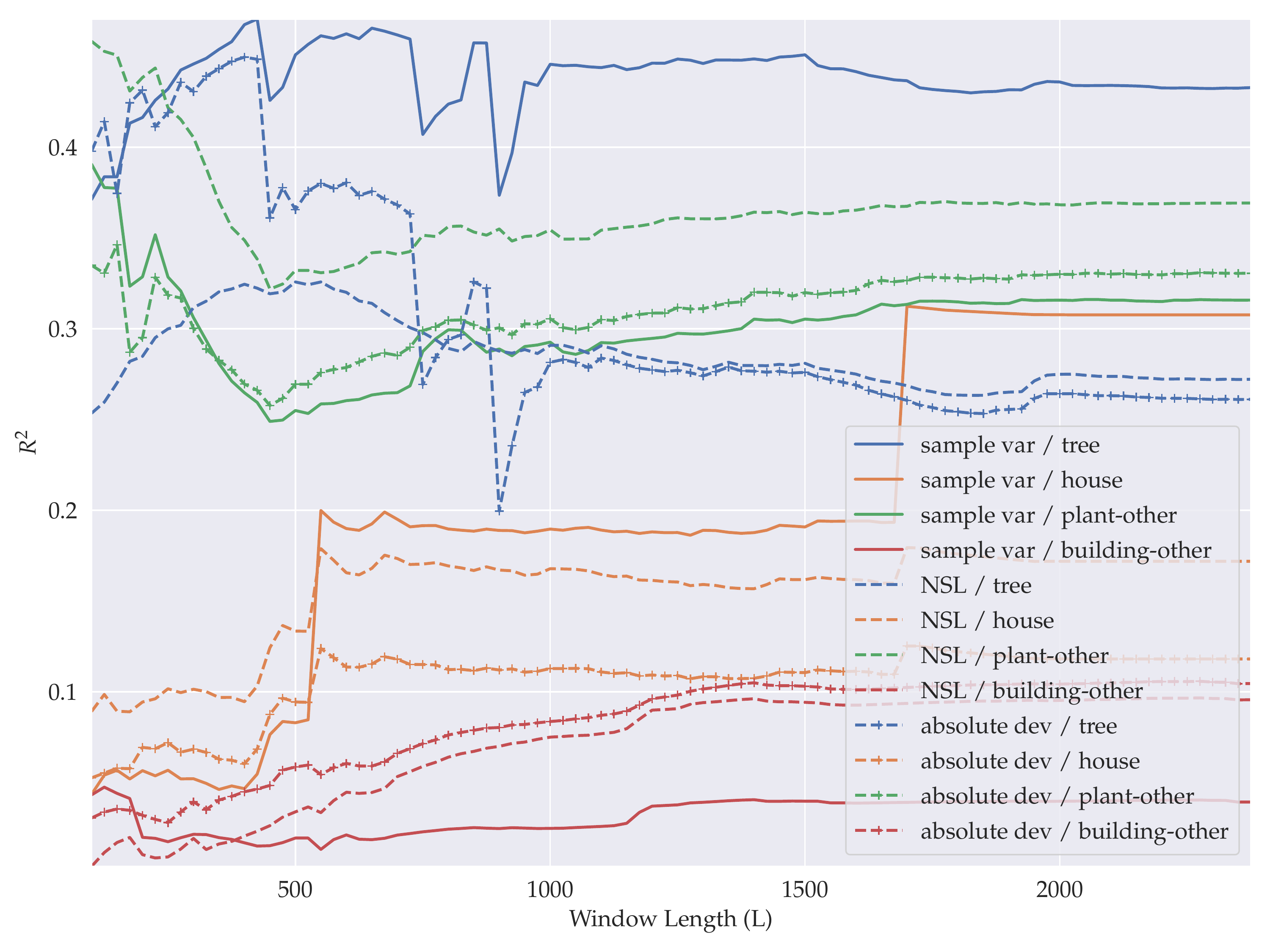}
	\caption{Impact of windowing on linear regression results: coefficient of determination, $R^2$, as a function of the window size, $L$, for the different object classes and metrics.}
	\label{fig:r2vswindowing}
\end{figure*}

\begin{figure*}[!ht]		
    \centering
	\includegraphics[width=0.7\linewidth]{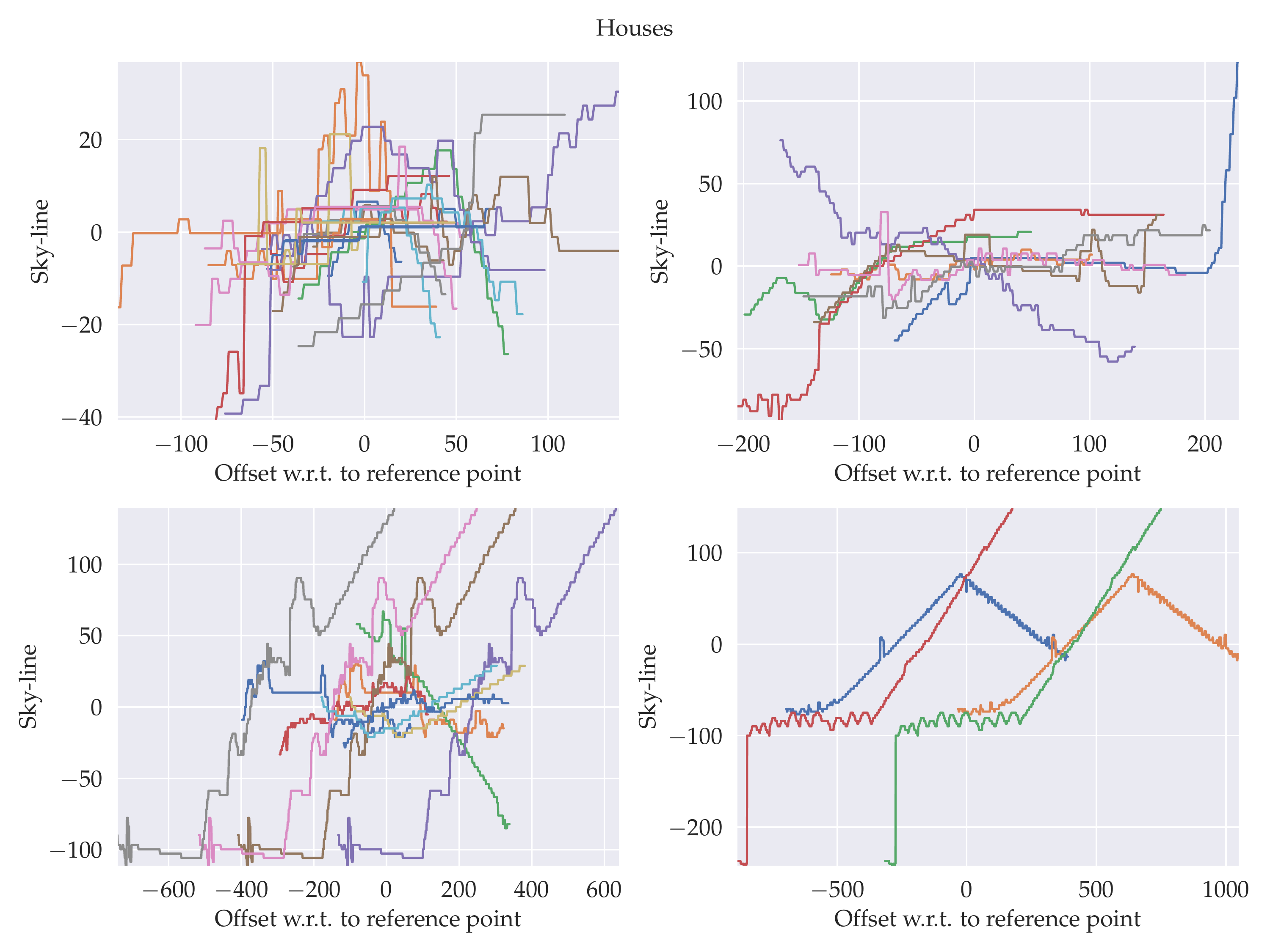}
	\caption{Skyline profiles of objects belonging to the ``Houses'' category. Profiles have been aligned with respect to their mean and horizontally centred with respect to the corresponding reference point. Repeated profiles correspond to different reference points. To improve clarity, the profiles have been divided in four subplots according to their width in pixels.}
	\label{fig:houseprofiles}
\end{figure*}

The analysis discussed above has been repeated by introducing windowing. In particular, linear regression has been performed by considering windowed metrics with a variable window length. The $R^2$ obtained for the different metrics and for the different object classes is provided in \fref{fig:r2vswindowing} as a function of the window length, $L$. For the ``Trees'' class, the sample variance led to an $R^2$ practically always above $0.4$. Some slight reductions are observed only for very low values of $L$, which suggests that a minimum number of pixels should be used for the computation of the sample variance. The maximum $R^2$ obtained for this case is around $0.47$ and is obtained for $L = 425$. Also the other metrics have a peak around $L = 425$: a progressive decrease is then observed.
For large values of $L$, the $R^2$ values reported in \fref{fig:r2valuescomp} are found. For values of $L$ lower than $800$, the \ac{NSL} is the metric with the lowest $R^2$ whereas, for larger values, it has a behaviour similar to that of the absolute deviation.\\
When the ``Other Plants'' class is considered, the best results are obtained when considering the \ac{NSL}. While it is difficult to determine the root cause of this effect, it is important to remind that only $33$ points belong to this class. Thus, this result needs to be confirmed by considering larger datasets. Also in this case, larger $R^2$ values are found when considering $L < 500$.\\
The ``Other Buildings'' class is characterized by the lowest $R^2$ values which are always lower then $0.105$. Windowing further reduces the $R^2$ value confirming that the variability of the sky-line profile of this class is only marginally affected by distance.\\
Finally, a behaviour strongly influenced by the window length, $L$, is found for the ``Houses'' class. In particular, two jumps in the $R^2$ value are found. These jumps occur for $L = 550$ and $L = 1700$. Note that windowing is applied on both sides of the corresponding reference point, $x_i$. If $x_i$ is close to one of the borders of the image, one side of the skyline profile won't be affected whereas the opposite one will be limited to $L/2$ pixels. For this reason, values of $L$ up to twice the image width should be considered.\\
The appearance of jumps in the $R^2$ curve is due to the presence of specific profiles with large vertical variations. Without windowing or for large values of $L$, these large vertical variations cause significant metric values, which are much larger then the bulk of the observations. These can be considered outliers that bias the regression process and lead to larger $R^2$ values. This fact is analysed in \fref{fig:houseprofiles} which shows the $43$ sky-line profiles belonging to the ``Houses'' classes. The profiles have been vertically centred with respect to their mean and horizontally aligned with respect to the corresponding reference point. In this way, profile values at $x = 0$ correspond to the reference points. Some profiles are repeated with a horizontal shift as they correspond to different reference points. As mentioned in \secref{sssec.wm}, without windowing, these profiles would lead to the same metric value for different distances. 
For clarity, the profiles have been divided in four subplots: profiles with width lower than $200$ pixels, between $200$ and $400$ pixels, between $400$ and $800$ pixels and for widths above $800$ pixels. Only few profiles are present in the last two categories shown in the bottom row of \fref{fig:houseprofiles}.
The second jump in \fref{fig:r2valuescomp} at $L = 1700$ is caused by the four profiles in the bottom right box of \fref{fig:houseprofiles}. For such large value of $L$, the large vertical jumps in these four profiles is captured and large sample variances are obtained for very short distances. These points can be clearly seen in the upper right box of \fref{fig:sample_var_reg}: they are almost coincident and practically touch the top point of the regression line. Without these four profiles the second jump in the $R^2$ value does not occur. If the four repeated profiles in the bottom left part of \fref{fig:houseprofiles} are removed an additional reduction of $R^2$ is observed. In this case, the $R^2$ remains below $0.2$. While it was not possible to identify the profiles causing the first jump, a similar phenomenon as the one described for the second jump is expected.\\
These results show that the sample variance, which is not a robust operator \cite{Huber2009}, can be severely affected by profiles such those identified in the bottom row of \fref{fig:houseprofiles}. Without such profiles, lower $R^2$ values are found also for the ``Houses'' class confirming the intuition that lower distance dependency should be expected for this object category. Windowing significantly reduces the impact of such profiles and low $R^2$ values are found for $L < 550$ pixels. Other metrics, such as the absolute variation, which is a robust operator, are less affected by these types of artefacts.\\
The results just discussed show the benefits of windowing. Moreover, they suggest that $L$ should be selected in the $[400-500]$ range.
\begin{figure*}[!ht]	
	\centering
	\includegraphics[width=0.7\linewidth]{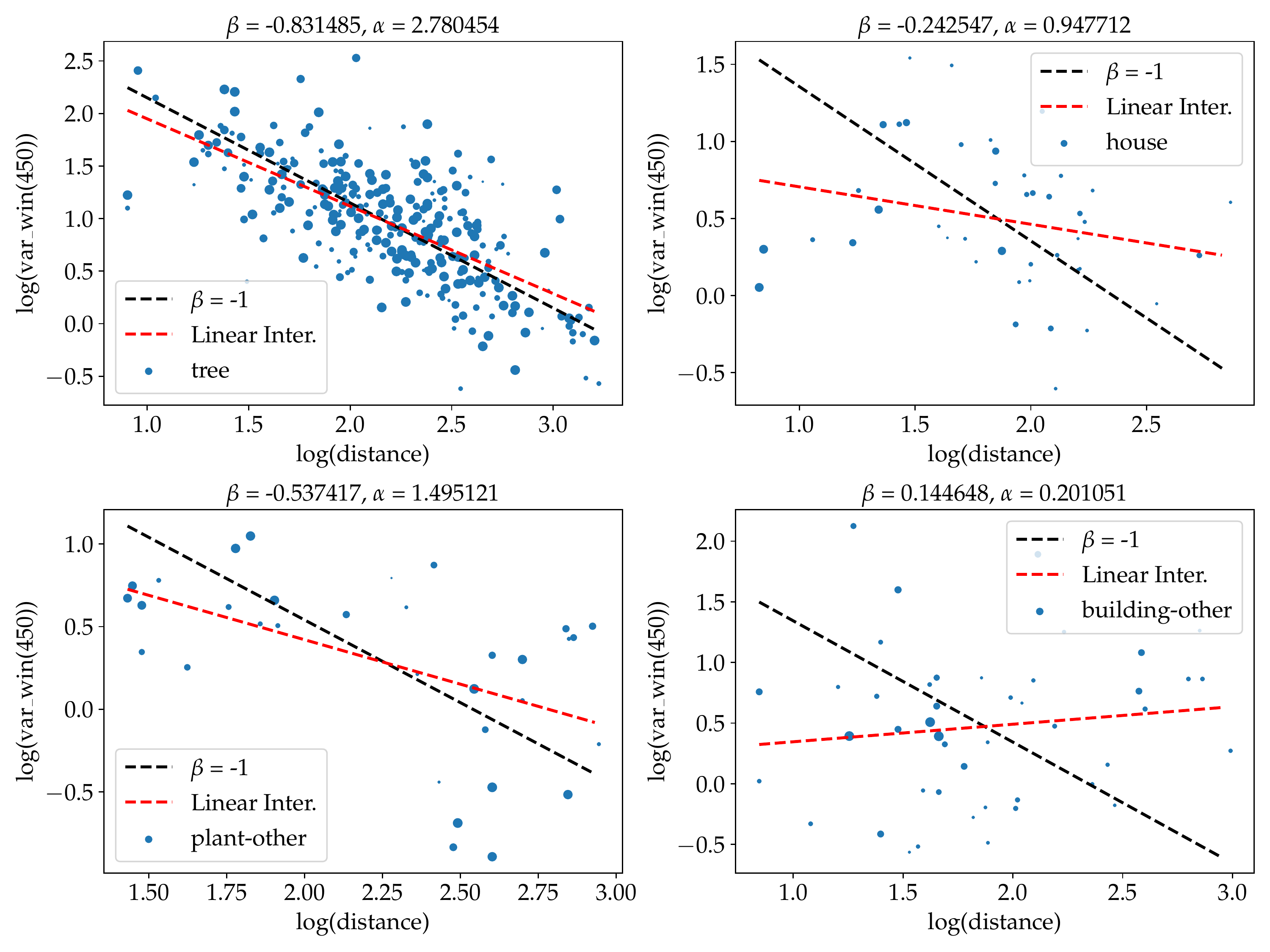}
	\caption{Scatter plots of the windowed sample variance with $L = 425$ as a function of the estimated distance for different object classes. Regression lines are also provided.}
	\label{fig:samplevarwin425reg}
\end{figure*}
In this respect the scatter plots of the windowed sample variance with $L = 425$ are provided in \fref{fig:samplevarwin425reg}. The ``Trees'' class scores an $R^2$ above $0.47$ and have a slope close to $\beta = -1$. Windowing also reduces the impact of outliers and the $R^2$ values for the ``Houses'' and ``Other Buildings'' are $0.055$ and $0.015$, respectively. This confirms the intuition that the variation in the skyline profiles of these classes do not show a dependency on distance.\\
This result suggests that this metric, applied to trees and natural objects, can be effectively used to infer information on the distance from the camera to the skyline.

\section{Discussion}
\label{sec.discussion}
In this paper, a dependency between the camera-to-object distance and the variability of the object profile was shown for certain classes of objects, such as trees and plants. The relationship between distance and variability metrics is an inverse power law and allows identifying the order of magnitude of the distance to the object. In turn, this information can be used to infer characteristics of the image and of the landscape.\\
While the present research was originally developed to estimate landscape openness, as determined by skyline objects and by their proximity to the camera, the results found have connections with recent developments in computer vision and in particular with depth estimation from a single image \cite{Saxena2008,Tung2019}. As an example, monocular depth estimation with supervised dense depth prediction per single image has been very successful with deep neural networks recently \citep{casser2019depth}. While in classic photogrammetry, information from two photographic images of the same landscape are needed to develop a scene stereographic model with accurate object distances \cite{Palmisano2010}, more recent research aims at the same goal using a single image, exploiting information embedded in the image objects. This is the same process adopted here where the variations of the skyline profiles have been investigated. For instance, Tung and Hwang \cite{Tung2019} developed a method to  generate a depth map from a single image that involved a sequence of blurring and deblurring on each point. Their work considered the relationship between the out-of-focus blurriness of a pixel and the distance of the related object in the scene.

Additional investigation is need to determine potential relationships between blurriness and level of details/variability of object profiles.\\
The method we developed in this paper has been validated considering objects with distances ranging from a few meters to some kilometers. At the larger distances, lack of detail in the signal will saturate the model. Distances have been evaluated using measurements from ortho-photos. Thus, the measured distances can be affected by errors due to: 1) inaccuracies in the location of the camera location, 2) inaccuracies in the location of the reference object, 3) limitations in the measurement process inherent to the capabilities of the graphical tools provided by QGIS, 4) by potential errors from the human operator and 5) by error of ortho-rectication including terrain and sensor geometries corrections. All these errors contributed to the unexplained components in the model assessed in the regression analysis performed in the paper. Future work will involve precise field measurements using tape meters and, for distant objects, precise locations from \ac{GNSS} receivers.\\
 
Finally, the metric developed here can become useful in various application contexts. 
For example, the metric could be used to quantify (perceptions of) landscape openness based on landscape photos, for the simultaneous object location on oblique (e.g. street-level imagery) and ortho-imagery, and may be valuable to gather in-situ data for Earth Observation. The first example relates to the concept of openness. In situations where $360$ degree photos are available, or as in the case of the \ac{LUCAS} photos dataset where photos are taken at each point in the four cardinal directions, the summarized distances to objects may be a proxy to determine how humans experience openness of the landscape. A second application context is in semi-automated cross-referencing of located objects with ortho-photos to confirm their location, for example monumental trees. Following further processing, the estimated distances also could be used as a proxy to estimate the extent of homogeneous land cover surfaces seen on the imagery that is bordered by trees, for example an agricultural field. Such information may be valuable to generate in-situ data for Earth Observation applications. The approach implemented not only involved the determination of the skyline signals, but semantic segmentation where different objects are identified. The analysis performed showed that different objects such as trees and buildings can be reliably classified and their profile extracted. Thus, the approach proposed can be adopted, for example, to detect woody vegetation in photos \cite{Bayr2019} and support applications such as monitoring of woody vegetation regrowth.  

\section{Conclusions}
\label{sec.conclu}
The visual detail of objects diminishes as they are farther away. In addition, natural objects, such as trees, have a fractal-like visual appearance. These key principles have been exploited in this work to investigate potential relationships between object distances and metrics quantifying the variability of their profiles. Semantic segmentation has been used to determine skyline profiles and determine the class of the underlying object. The profiles obtained have then be used to compute variability metrics, which exhibit a dependency with object distances. Among the metrics investigated the sample variance is the one exhibiting the strongest distance dependency. Regression analysis has been performed and, in the best case, a $R^2 = 0.47$ was found for the class of trees. This implies that a significant component of the variations in sample variance can be explained by a power law model depending on distance. The work presented demonstrates that skyline profile variations can be effectively used, for certain classes of objects, to approximately determine distances from the camera. 

\section{Acknowledgements}
We gratefully acknowledge the support of this research by the JRC Exploratory Research program through the Rural Refocus project (31280).

\bibliography{references}

\end{document}